\PassOptionsToPackage{table,dvipsnames}{xcolor}
\documentclass[]{selfevolagent}

\usepackage{microtype}
\usepackage{amsfonts}
\usepackage{amsmath}
\usepackage{amsthm}
\usepackage{xcolor}
\usepackage{amssymb}
\usepackage{graphicx}     
\usepackage{booktabs}     
\usepackage{tabularx}
\usepackage{longtable}
\usepackage{textcomp}
\usepackage{xltabular}
\usepackage{makecell}
\usepackage{multirow}
\usepackage{wrapfig}
\usepackage{float}
\usepackage{subcaption}
\usepackage{caption}
\usepackage{enumitem}     
\usepackage{pifont}       
\usepackage{url}
\usepackage[most]{tcolorbox} 
\usepackage{fontawesome5}    
\usepackage{algpseudocode}
\usepackage[linesnumbered,lined,boxed,commentsnumbered,ruled,longend]{algorithm2e}
\usepackage[T1]{fontenc}
\usepackage{siunitx}
\usepackage{newtxtext,newtxmath}

\DeclareCaptionFormat{coloncaption}{#1#2#3}
\usepackage{tikz}
\usepackage[edges]{forest}
\usetikzlibrary{
  arrows.meta,
  positioning,
  calc,
  shapes.symbols,
  shapes.geometric,
  shapes.misc
}

\theoremstyle{plain}

\theoremstyle{definition}

\theoremstyle{remark}

\definecolor{themecolor}{HTML}{37D2A6}        
\definecolor{themecolor_light}{HTML}{9BE9D3}
\definecolor{themecolor_lighter}{HTML}{CDF4E9}

\definecolor{contributionblue}{RGB}{0,0,120}

\newcommand{\ghlink}[1]{\faIcon{github}\,\href{#1}{GitHub}}
\newcommand{\weblink}[1]{\faIcon{globe}\,\href{#1}{Website}}

\title{AREX: Towards a Recursively Self-Improving Agent for Deep Research}

\author{AREX Team}

\affiliation{
\vspace{0.5em}
Beijing Academy of Artificial Intelligence (BAAI)
}
\vspace{0.5em}

\abstract{

Deep research requires agents to find answers that jointly satisfy multiple constraints. 
Discovering such answers is costly, whereas verifying a candidate can often be decomposed into tractable constraint-wise checks. 
This discovery--verification asymmetry suggests that a research agent should do more than simply search longer: it should recursively improve its current answer by verifying intermediate results and using the partially verified state to guide subsequent refinement.
We introduce \textsc{AREX}, a family of \emph{Recursively Self-Improving (RSI)} deep research agents. 
\textsc{AREX} alternates between an \textit{inner research loop} that gathers evidence and constructs a provisional answer, and an \textit{outer self-improvement loop} that audits the answer constraint-wise, identifies unresolved claims, and launches targeted follow-up research. 
To sustain RSI over long horizons, \textsc{AREX} learns an autonomous context-update tool that compresses growing interaction history into a compact improvement state preserving verified evidence and unresolved constraints, without relying on an external model. 
We train \textsc{AREX} on verified synthetic tasks and high-quality trajectories through agentic mid-training and long-horizon reinforcement learning. 
To mitigate sparse final rewards during long horizon learning, we emphasize key steps where decisive evidence is acquired or erroneous research directions are corrected. 
We instantiate a dense 4B model and a 122B-A10B Mixture-of-Experts model. Across BrowseComp, WideSearch, DeepSearchQA, Humanity's Last Exam (HLE), and other reasoning and tool-use benchmarks, \textsc{AREX} substantially outperforms comparable-scale baselines and remains competitive with models using substantially more activated parameters.

}

\metadata[{\textcolor{RoyalBlue}{\faIcon{globe}}\ App}]{\url{https://arex-research.com}}
\metadata[{\raisebox{-0.2ex}{\includegraphics[height=1em]{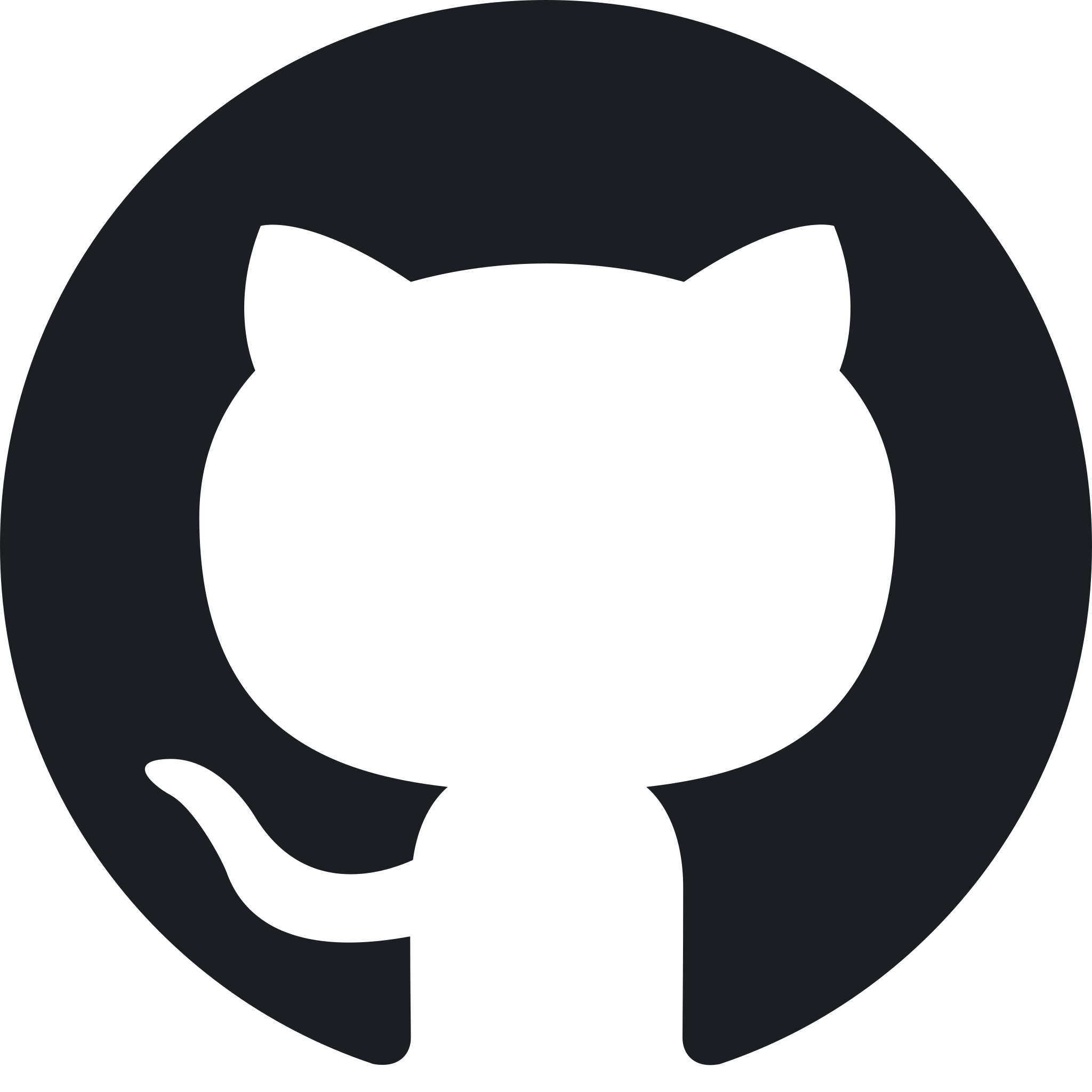}}\ Homepage}]{\url{https://vectorspacelab.github.io/arex-model/}}
\metadata[{\raisebox{-0.2ex}{\includegraphics[height=1em]{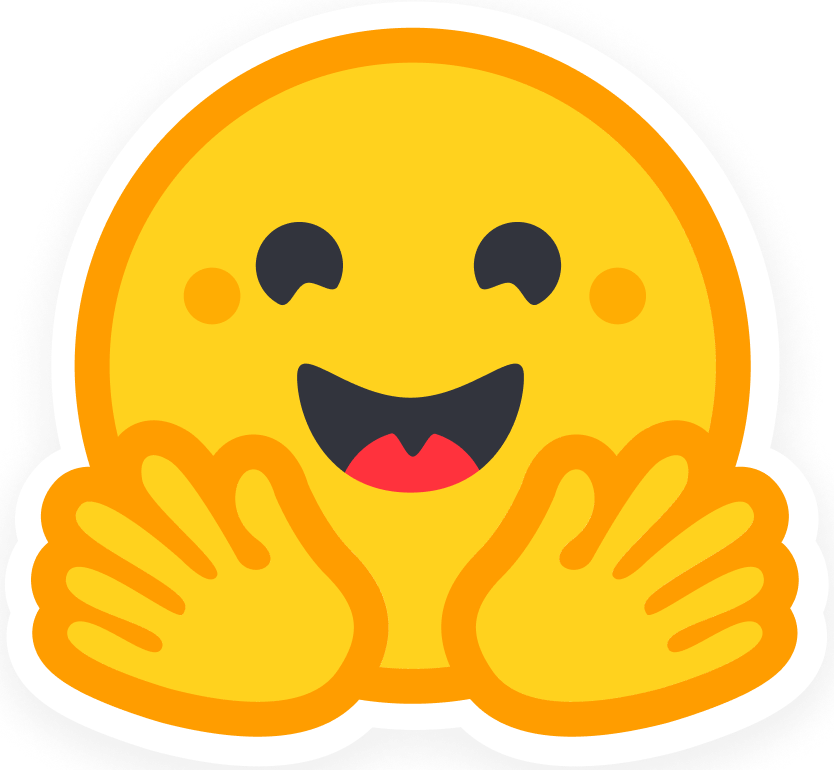}}\ Models}]{\url{https://huggingface.co/collections/BAAI/arex}}

\begin{document}
\maketitle

\begin{figure}[!ht]
    \centering
    \includegraphics[width=\linewidth]{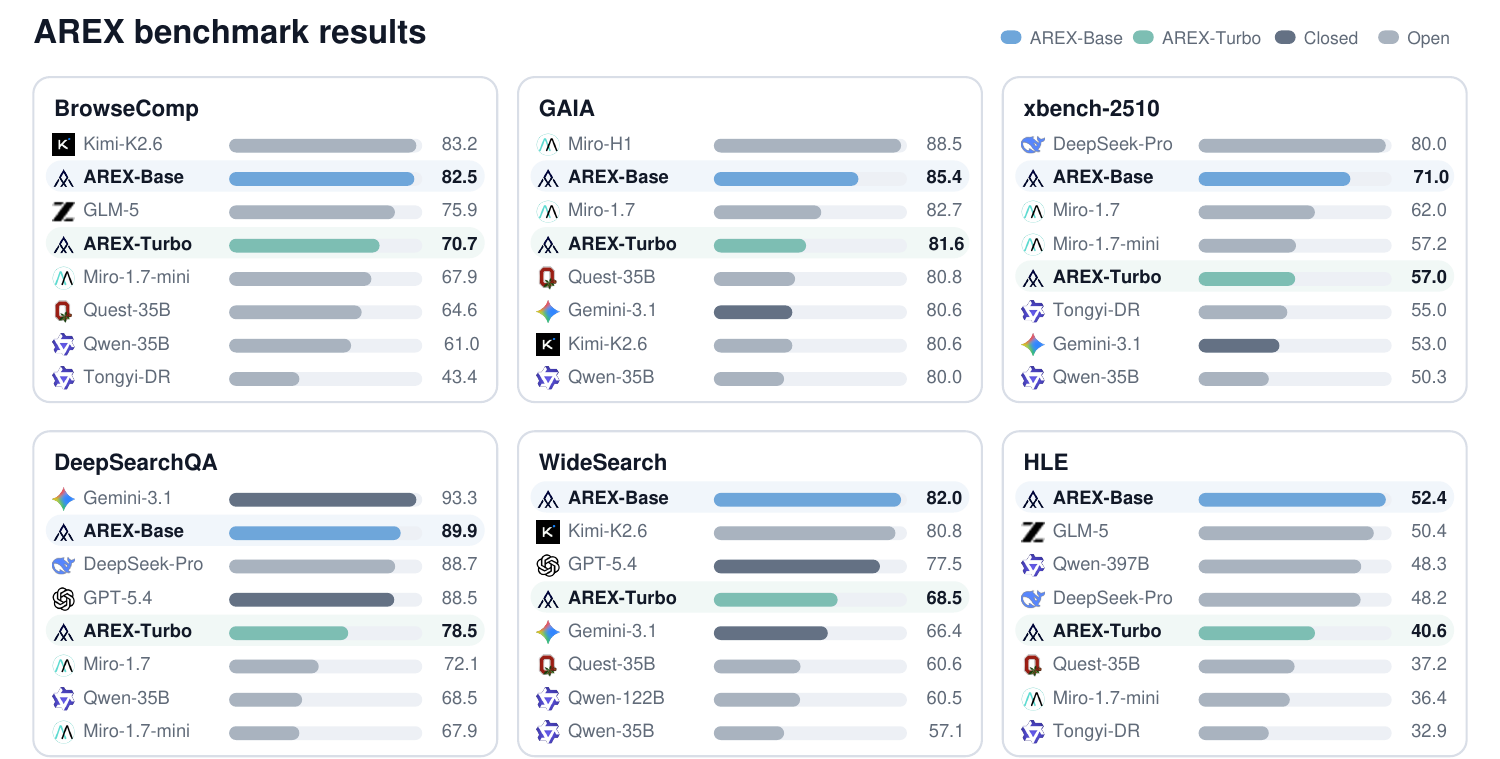}
    \caption{Benchmark performance of AREX}
    \label{fig:placeholder}
\end{figure}

\section{Introduction}
\label{sec:intro}


Deep research is challenging not only because relevant evidence is difficult to locate, but because a valid answer must often satisfy multiple coupled constraints simultaneously. 
An agent must discover viable candidates, integrate distributed and potentially conflicting evidence, and verify that each required condition is adequately supported~\citep{react,webgpt,deepresearch_survey}. 
Many existing systems address this challenge by extending a single search trajectory with additional reasoning, tool interactions, or context~\citep{deepresearch,searchr1,webdancer,beyond_ten_turns}. Although greater inference-time computation can broaden exploration, it does not guarantee systematic progress: early errors may persist, exhausted directions may be revisited, and partially valid candidates may be accepted prematurely. The key challenge is therefore not merely to search longer, but to identify which constraints remain unresolved and use that diagnosis to formulate a more targeted next research problem.

We observe that deep research tasks exhibit a fundamental \emph{discovery--verification asymmetry}. Discovering an answer that jointly satisfies all constraints is costly because it requires navigating a large and sparsely informative search space, whereas evaluating a proposed candidate can often be decomposed into substantially simpler constraint-wise checks. 
Such verification reveals not only whether the current answer is correct, but also which claims are supported, which remain unresolved, and where the available evidence conflicts. 
Existing work leverages verification either to rank completed candidate trajectories or to refine decisions within an ongoing trajectory~\citep{zeng2025pushing,team2026mirothinker}. 
Our key insight is that verification can also define the transition between research rounds. By converting a provisional answer into a partially verified state, the agent can preserve established progress, isolate the remaining uncertainty, and formulate a more targeted next research problem. 

We introduce \textsc{AREX}, a family of \emph{Recursively Self-Improving (RSI)} deep research agents that repeatedly convert partially verified solutions into better-targeted research problems. 
\textsc{AREX} alternates between an inner \emph{research loop}, which gathers evidence, evaluates candidates, and constructs a provisional answer, and an outer \emph{self-improvement loop}, which audits the answer against the task constraints and directs subsequent research toward unresolved or weakly supported claims.
Verified progress is preserved across rounds, while constraint-level belief estimates govern both continuation and termination: uncertain conditions trigger targeted follow-up research, and the process terminates once the required claims are sufficiently supported. 
In this way, verification becomes an active control signal that recursively refines the agent's research state and solution, rather than a final filter applied after search. 

Sustaining this recursion over long horizons requires the agent to maintain a concise and actionable research state. 
During the inner research loop, the interaction history continually accumulates verified evidence, failed queries, speculative hypotheses, duplicated observations, and outdated plans. 
Retaining the full history can distract subsequent reasoning, whereas indiscriminate truncation may discard evidence needed for later verification~\citep{memgpt,hiagent,mem1,resum,supo,memory_as_action}. 
\textsc{AREX} therefore learns to autonomously invoke a dedicated context-update tool during research, converting its own interaction history into a compact \emph{improvement state}. This state preserves verified evidence and citations, records constraint-satisfaction status, highlights unresolved information gaps, and specifies the next research plan. 
Unlike generic summarization or compression performed by an external model, the update is produced by \textsc{AREX} itself and is organized around its current research objective. 
This autonomous context management keeps the compressed state aligned with the agent's evolving beliefs and future actions.

We train \textsc{AREX} on verified synthetic tasks and high-quality trajectories through supervised capability acquisition, agentic mid-training, and reinforcement learning for long-horizon research. 
These stages progressively teach the model to search, use tools, construct provisional answers, verify individual constraints, update context, and decide when to continue or terminate. 
Long trajectories also create a sparse credit-assignment problem: final rewards do not reveal which intermediate actions produced decisive progress. 
We therefore identify and increase training exposure to \emph{key steps}, such as steps where key evidence is acquired, contradictions are resolved, or an incorrect research direction is repaired~\citep{lets_verify,math_shepherd,ragen,turn_credit}. 

We instantiate \textsc{AREX} as a dense \textbf{4B} model (Turbo) and a Mixture-of-Experts model with \textbf{122B total parameters} and \textbf{10B activated parameters} (Base). 
We evaluate them on benchmarks spanning deep search, wide search, multi-constraint information seeking, and reasoning with tools, including BrowseComp~\citep{browsecomp}, WideSearch~\citep{widesearch}, DeepSearchQA~\citep{deepsearchqa}, Humanity's Last Exam (with Tools)~\citep{hle}, GAIA~\citep{gaia}, and xbench-DeepSearch-2510~\citep{xbench_deepsearch_2510}. 
Across these settings, \textsc{AREX} substantially outperforms comparable-scale baselines and remains competitive with models using substantially more activated parameters, demonstrating that recursively improving the research state offers an effective path toward capable and efficient deep research agents.

Our main contributions are summarized as follows:
\begin{itemize}
\item We formulate multi-constraint deep research as a \emph{Recursively Self-Improving} process motivated by discovery--verification asymmetry, in which partially verified solutions are recursively converted into better-targeted research problems.
\item We introduce \textsc{AREX}, which combines a research loop with a constraint-wise self-improvement loop. Belief estimates govern targeted continuation and evidence-aware termination, while a learned context-update tool maintains a compact improvement state over long horizons.
\item We develop a multi-stage training framework based on verified synthetic tasks and high-quality trajectories, together with critical-interval exposure for improving credit assignment in long research trajectories.
\item We develop dense 4B and 122B-A10B MoE variants and demonstrate consistent gains over comparable-scale baselines across deep-research, wide-search, reasoning, and tool-use benchmarks.
\end{itemize}

\begin{figure}[!t]
\vspace{-3em}
    \centering
    \includegraphics[width=\linewidth]{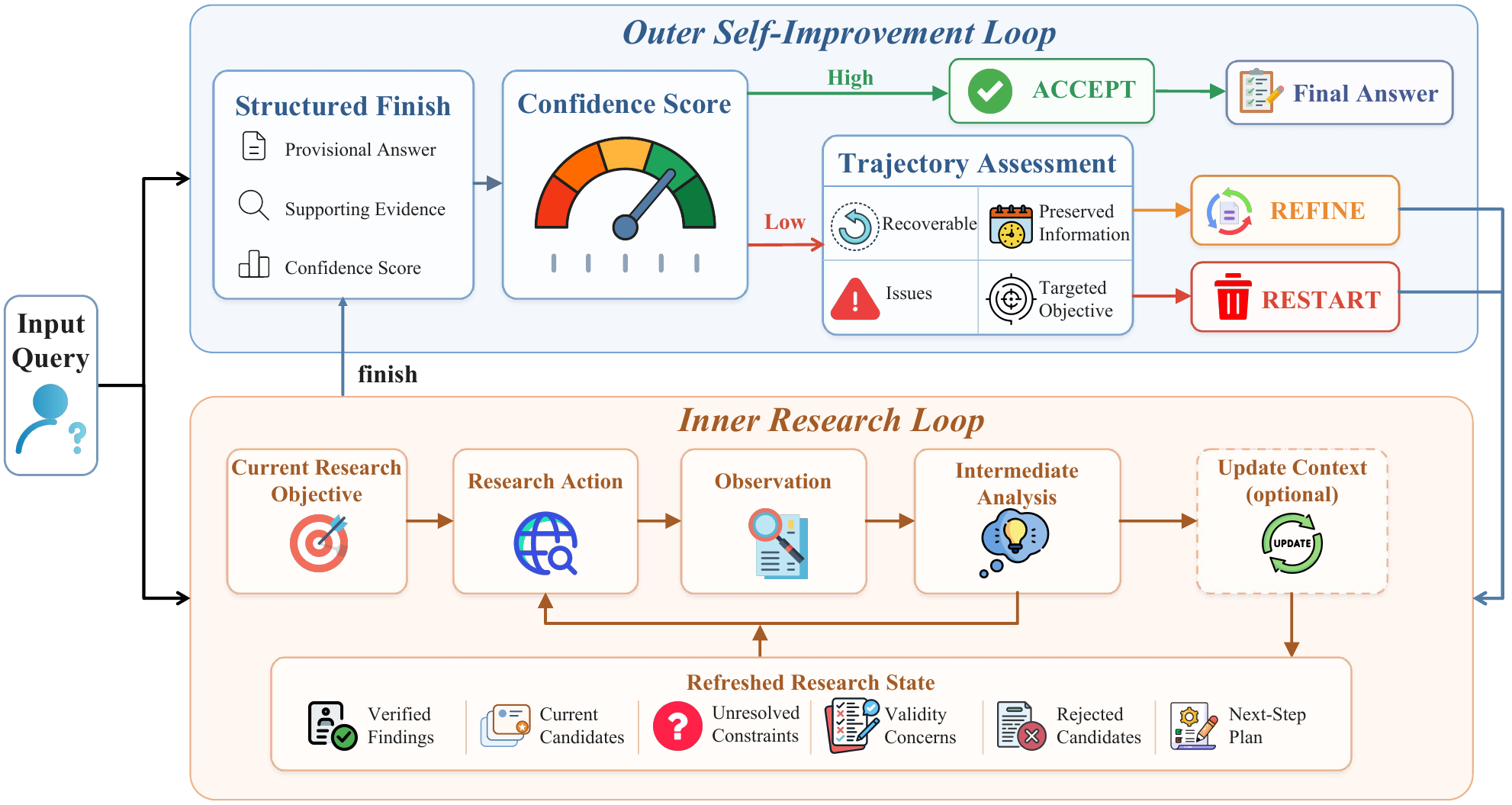}
    \caption{\textsc{AREX}'s recursive self-improvement framework. The inner research loop maintains the research state and externalizes a provisional answer with supporting evidence and confidence, while the outer self-improvement loop accepts, refines, or restarts based on confidence and trajectory assessment.}
    \label{fig:method}
\end{figure}
\vspace{-0.5em}

\section{Recursive Self-Improvement}

\subsection{Overall Framework}

We adopt Qwen3.5-4B~\citep{qwen35} as the backbone model for \textsc{AREX-Turbo} and Qwen3.5-122B-A10B~\citep{qwen35} for \textsc{AREX-Base}.
As illustrated in Figure~\ref{fig:method}, \textsc{AREX} implements deep research through a hierarchical, bi-level recursive self-improvement process, comprising an inner research loop and an outer self-improvement loop. Given an input query $x$, it first derives a research objective $q^{(1)}$. The \emph{inner research loop} executes research actions, integrates the retrieved evidence, and updates the current answer until the objective is sufficiently addressed. It then outputs a provisional answer together with supporting evidence and an answer-level confidence score.

The \emph{outer self-improvement loop} evaluates the provisional result using its confidence score. If the score exceeds a predefined threshold, the answer is accepted. Otherwise, the loop assesses whether the current research trajectory is recoverable: useful findings are preserved and converted into a targeted objective for refinement, while noisy or uninformative trajectories trigger a restart from the original problem.

\subsection{Inner Research Loop}

The inner research loop incrementally constructs an answer by analyzing the current research objective, invoking search or browsing tools, integrating the returned evidence, and determining the next research step. In the first recursive round, the objective $q^{(1)}$ is derived from the original problem $x$. In subsequent rounds, the outer loop may provide a targeted objective $q^{(k)}$, such as verifying an unsupported constraint, resolving conflicting evidence, checking temporal validity, or exploring alternatives after a candidate has been invalidated.

At step $t$ of recursive round $k$, the interaction trajectory $h_t^{(k)}$ is defined as
\begin{equation}
h_t^{(k)}
=
\left[
\left(
m_i^{(k)},
a_i^{(k)},
o_i^{(k)}
\right)
\right]_{i=1}^{t},
\end{equation}
where $m_i^{(k)}$ denotes the model's intermediate analysis, $a_i^{(k)}$ a research action, and $o_i^{(k)}$ the corresponding observation.

The trajectory $h_t^{(k)}$ serves as the model's working research state within recursive round $k$. Based on the original problem $x$, the current research objective $q^{(k)}$, and the accumulated trajectory $h_t^{(k)}$, the model identifies unresolved constraints and produces an intermediate analysis together with a corresponding research action:
\begin{equation}
\left(
m_{t+1}^{(k)},
a_{t+1}^{(k)}
\right)
=
\pi_{\theta}
\left(
x,
q^{(k)},
h_t^{(k)}
\right),
\qquad
o_{t+1}^{(k)}
=
\mathcal{T}
\left(
a_{t+1}^{(k)}
\right),
\end{equation}
where $\pi_{\theta}$ denotes the research policy parameterized by $\theta$, and $\mathcal{T}$ denotes the external research environment, including search and browsing tools. The resulting interaction is appended to the trajectory:
\begin{equation}
h_{t+1}^{(k)}
=
h_t^{(k)}
\oplus
\left(
m_{t+1}^{(k)},
a_{t+1}^{(k)},
o_{t+1}^{(k)}
\right),
\end{equation}
where $\oplus$ denotes chronological concatenation.

The research policy adapts as new evidence is accumulated in the trajectory. Supporting observations guide the model toward remaining constraints, whereas contradictory observations may invalidate current candidates and redirect the investigation toward alternatives. When sources conflict, the model seeks evidence with stronger authority, more direct provenance, or greater temporal relevance. If no plausible candidate remains, it may broaden the search space, decompose the objective, or reformulate the search query.

During long-horizon research, the model may invoke \texttt{update\_context} to consolidate the accumulated trajectory into a more compact research state, allowing subsequent steps to operate on the refreshed context while preserving salient evidence and unresolved constraints. The loop terminates when the current objective has been sufficiently investigated or further search is unlikely to provide substantial benefit. It then invokes \texttt{finish} to externalize a provisional answer, its supporting evidence, and an answer-level confidence score. The following two mechanisms describe, respectively, how the trajectory is refreshed during execution and how the terminal trajectory is converted into an answer-level representation for outer-loop evaluation.

\subsubsection{Autonomous Context Updating}

As research proceeds, the trajectory accumulates search results, intermediate conclusions, rejected candidates, conflicting findings, and evolving plans. Retaining the complete trajectory introduces redundancy and may obscure information relevant to subsequent decisions.

Existing approaches often manage context using fixed heuristics, such as discarding tool responses according to predefined rules~\citep{team2026mirothinker} or triggering summarization at a fixed token threshold~\citep{liu2025deepseek}. While these strategies reduce context length, they treat context management mainly as a budget-control problem rather than a research-state maintenance problem. As a result, they may remove or dilute decision-relevant information, including source provenance, negative evidence, unresolved constraints, conflicting findings, and the reasons why certain candidates have been rejected.

This limitation becomes more severe in long-horizon research, where the next useful action depends on semantic progress rather than message position or token count. The model must track which constraints have been satisfied, which hypotheses remain viable, and which uncertainties should guide further search. Fixed heuristics are not aligned with these research-state transitions, and may therefore cause the model to revisit invalidated candidates, rediscover prior conclusions, or lose information needed for later refinement.

To address this issue, \textsc{AREX} provides an explicit \texttt{update\_context} tool. Given the accumulated trajectory $h_t^{(k)}$, the tool constructs a refreshed research state:
\begin{equation}
z_t^{(k)}
=
f_{\theta}
\left(
h_t^{(k)}
\right).
\end{equation}
The refreshed state preserves verified findings and their source identifiers, current candidates, unresolved constraints, validity concerns, rejected candidates, and the next-step plan. Redundant observations, superseded conclusions, and obsolete plans are removed.

After \texttt{update\_context} is invoked, the model no longer needs to condition on the entire trajectory $h_t^{(k)}$. We denote the effective context available to the model at step $t$ by $\bar{h}_t^{(k)}$. If the most recent \texttt{update\_context} invocation occurs at step $\tau \leq t$, then
\begin{equation}
\bar{h}_t^{(k)}
=
z_\tau^{(k)}
\oplus
\left[
\left(
m_i^{(k)},
a_i^{(k)},
o_i^{(k)}
\right)
\right]_{i=\tau+1}^{t}.
\end{equation}
If no context update has occurred, the effective context is simply the complete trajectory:
\begin{equation}
\bar{h}_t^{(k)} = h_t^{(k)} .
\end{equation}
Subsequent actions are generated from the effective context:
\begin{equation}
\left(
m_{t+1}^{(k)},
a_{t+1}^{(k)}
\right)
=
\pi_{\theta}
\left(
x,
q^{(k)},
\bar{h}_t^{(k)}
\right).
\end{equation}

The model autonomously determines when to invoke \texttt{update\_context}, for example after resolving a meaningful subproblem, eliminating a major candidate, reconciling conflicting evidence, or changing its research plan. The tool may therefore be invoked multiple times within an inner loop or not at all.

Rather than generic summarization, autonomous context updating performs trajectory consolidation and research-state refreshing. It enables long-horizon research without requiring the model to repeatedly reconstruct progress from the complete interaction history and allows useful findings and unresolved constraints to be reused across recursive rounds.

\subsubsection{Structured Answer Externalization}

Autonomous context updating maintains the trajectory during research, but the resulting state is not itself an answer. 
Once the current objective has been sufficiently investigated, the model invokes a structured \texttt{finish} interface.

Let $\bar{h}_{T_k}^{(k)}$ denote the terminal effective context produced by the inner loop, where $T_k$ is the final research step of recursive round $k$. At recursive round $k$, the inner-loop output is
\begin{equation}
r^{(k)}
=
F_{\theta}
\left(
\bar{h}_{T_k}^{(k)}
\right)
=
\left(
y^{(k)},
\mathcal{E}^{(k)},
s^{(k)}
\right),
\end{equation}
where $T_k$ is the final step, $y^{(k)}$ is the provisional answer to the original problem, $\mathcal{E}^{(k)}$ contains supporting evidence and document identifiers, and $s^{(k)} \in [0,100]$ is an answer-level confidence score.

The confidence score reflects the estimated completeness, consistency, provenance, and temporal validity of the answer. Unlike the full trajectory, which may include discarded directions, obsolete plans, and raw tool responses, $r^{(k)}$ retains only the answer-level information needed for outer-loop evaluation.

Invoking \texttt{finish} terminates the current inner loop rather than the entire recursive process. The outer loop evaluates the provisional answer, evidence, and confidence score against the original problem, and either accepts the answer or formulates a targeted objective for another research round.

\subsection{Outer Self-Improvement Loop}

At recursive round $k$, the outer self-improvement loop receives the structured result $r^{(k)}$ together with the effective terminal context $\bar{h}_{T_k}^{(k)}$. It then decides whether to terminate the recursive process, refine the current trajectory, or restart the investigation.

The outer loop uses the confidence score as a compact summary of the evidence-grounded belief state constructed by the inner loop, including verified findings, conflicting evidence, and unresolved constraints. 
Let $\tau$ denote the confidence threshold. If $s^{(k)} \geq \tau$, the current answer is accepted and returned as the final answer. 
Otherwise, the model evaluates whether the current research trajectory contains useful information that can support further improvement.

For a low-confidence result, the trajectory assessment is defined as
\begin{equation}
g^{(k)}
=
G_{\theta}
\left(
x,
r^{(k)},
\bar{h}_{T_k}^{(k)}
\right)
=
\left(
v^{(k)},
\mathcal{P}^{(k)},
\mathcal{I}^{(k)},
q^{(k+1)}
\right),
\end{equation}
where $v^{(k)} \in \{0,1\}$ indicates whether the trajectory is recoverable, $\mathcal{P}^{(k)}$ contains information that should be preserved, $\mathcal{I}^{(k)}$ contains issues that require further investigation or revision, and $q^{(k+1)}$ is the research objective for the next recursive round.

The complete decision rule is
\begin{equation}
d^{(k)}
=
\begin{cases}
\textsc{Accept},
&
s^{(k)} \geq \tau,
\\[3pt]
\textsc{Refine},
&
s^{(k)} < \tau
\ \land\
v^{(k)} = 1,
\\[3pt]
\textsc{Restart},
&
s^{(k)} < \tau
\ \land\
v^{(k)} = 0.
\end{cases}
\end{equation}

When \textsc{Refine} is selected, the current trajectory is considered to contain meaningful progress. The outer loop preserves reliable findings and reusable evidence in $\mathcal{P}^{(k)}$, identifies the remaining problems in $\mathcal{I}^{(k)}$, and converts them into a targeted objective $q^{(k+1)}$. The next recursive round is initialized from a refreshed state:
\begin{equation}
h_{0}^{(k+1)}
=
\operatorname{Refresh}
\left(
\bar{h}_{T_k}^{(k)},
\mathcal{P}^{(k)},
\mathcal{I}^{(k)}
\right).
\end{equation}
This allows the next round to reuse useful progress while focusing on the parts of the answer that remain uncertain or incomplete.

When \textsc{Restart} is selected, the model determines that the current trajectory is too noisy, misleading, or uninformative to support further refinement. The accumulated trajectory is therefore discarded, and the next recursive round is initialized solely from the original problem:
\begin{equation}
h_{0}^{(k+1)}
=
\operatorname{Init}(x).
\end{equation}
The inner loop then begins a new investigation without inheriting information from the previous trajectory.

After each recursive round, the outer loop repeats the same confidence-based decision procedure. The process is bounded by a predefined maximum number of rounds. If no answer reaches the confidence threshold, the system returns the completed answer with the highest confidence score.

\section{Training Data Construction}

To enable \textsc{AREX} to perform recursive research, we construct a specialized training dataset containing challenging research tasks, multi-step investigation trajectories, and evidence-grounded solutions. The pipeline consists of two stages: recursive research task synthesis and teacher trajectory collection with quality control. The first stage generates verifiable  problem that require deep research, while the second collects high-quality trajectories from strong teacher models operating in the same research environment as \textsc{AREX}.

\subsection{Recursive Research Task Synthesis}

The goal of task synthesis is to construct problems that require iterative information gathering, multi-source evidence integration, intermediate hypothesis verification, and adaptive research planning. We consider three task categories: browse-intensive problems requiring information synthesis across sources, reasoning-intensive problems involving multi-step planning or deduction, and scientific literature problems requiring the integration of academic papers.

For each category, human experts define templates specifying the answer format, available sources, reasoning requirements, and verification criteria. Concrete instances are then generated from real-world sources, including web pages, scientific literature, structured knowledge bases, and public repositories.

We first extract a target entity or solution $y$ as the latent answer and identify a set of associated constraints:
\begin{equation}
\mathcal{C}(y)=\{c_1,c_2,\ldots,c_n\},
\end{equation}
where each constraint represents a verifiable research objective. These constraints may describe temporal relations, numerical properties, entity relations, technical attributes, or evidence requirements. Their combination forces the model to gather and reason over information from multiple sources rather than rely on keyword matching.

To avoid simple retrieval tasks, we abstract and transform the constraints into indirect descriptions requiring multi-hop search and reasoning. The final query is generated as
\begin{equation}
x=f(y,\mathcal{C}'),
\end{equation}
where $\mathcal{C}'$ denotes the transformed constraints. A valid task must satisfy three conditions: the answer cannot be inferred directly from the query, every constraint is verifiable from available evidence, and the joint constraints uniquely identify the answer.

We then automatically verify correctness, uniqueness, evidence availability, and difficulty. Tasks with ambiguous answers, inconsistent constraints, or insufficient evidence are removed. Independent research rollouts are further used to discard tasks that can be solved through shallow retrieval or remain unsolved after extensive investigation.

The resulting task dataset is
\begin{equation}
\mathcal{D}_{\mathrm{task}}
=
\{(x,y\},
\end{equation}
where each instance contains a research problem and its corresponding answer.

\subsection{Teacher Trajectory Collection and Quality Control}

For each synthesized task, strong teacher models interact with the same tools and research environment used by \textsc{AREX}. Given a task $x$, a teacher model samples a research trajectory:
\begin{equation}
\tau_i
\sim
\pi_{\mathrm{teacher}}(\tau\mid x).
\end{equation}
Each trajectory records model actions, tool calls, retrieved observations, intermediate analysis, and the final structured answer. Compared with standard input-output supervision, these trajectories expose the intermediate behaviors required for long-horizon research.

Because raw trajectories may contain ineffective exploration, reasoning errors, or unsupported conclusions, we apply several quality-control procedures. First, we retain only trajectories that exhibit meaningful iterative investigation, coherent research-state maintenance, and adaptation to newly acquired evidence. Trajectories that guess the answer directly, ignore observations, or fail to revise incorrect hypotheses are removed.

Second, we verify the validity of tool interactions. Trajectories containing invalid or unreliable tool executions, ignored observations, or unsupported references are discarded.

Third, we apply evidence-grounded answer filtering. The final answer must be reconstructable from the collected evidence. We remove trajectories that answer before gathering sufficient evidence, contain unsupported claims, contradict retrieved information, or rely on unjustified assumptions.

Finally, each trajectory terminates with an answer, supporting evidence, and an answer-level confidence score $s_{\mathrm{conf}}$. Trajectories satisfying
\begin{equation}
s_{\mathrm{conf}} < \tau_{\mathrm{conf}}
\end{equation}
are removed, where $\tau_{\mathrm{conf}}$ is a predefined confidence threshold.

After filtering, the final trajectory dataset is
\begin{equation}
\mathcal{D}_{\mathrm{traj}}
=
\{(x,\tau)\mid V(x,\tau)=1\},
\end{equation}
where $V(x,\tau)$ denotes the overall validity criterion. The resulting data supervise iterative evidence acquisition, research-state maintenance, evidence-grounded answer construction, and adaptive continuation, enabling \textsc{AREX} to operate effectively within its recursive double-loop framework.

\section{Training Pipeline}

\subsection{Multi-stage Agentic Mid-training}
\label{sec:mid-train}

Long-horizon deep research requires a combination of heterogeneous capabilities, including tool invocation, web navigation, evidence acquisition, expert-level reasoning, context tracking, and answer synthesis. 
Directly training the model on a single mixed distribution may lead to unstable learning dynamics and interference across different capability types. 
We therefore adopt a multi-stage agentic mid-training recipe that progressively improves the model's research ability.

The agentic mid-training process consists of two major stages. 
First, we train the model on browse-intensive research tasks to establish fundamental tool-use and evidence acquisition capabilities, and then train the model on expert reasoning tasks to strengthen long-form thinking, hypothesis comparison, and difficult problem solving. 
Second, we perform mixed-capability consolidation by replaying selected key steps from browse-intensive long-horizon trajectories and incorporating capability-expanding tasks, including complex academic paper research and challenging knowledge-intensive reasoning tasks. 
This stage also exposes the model to verification-driven research transitions, where a provisional answer is audited against task constraints, verified evidence is preserved in the research state, unresolved conditions are identified, and the next targeted research problem is formulated. This stage reinforces the model's core deep research capabilities while extending its reasoning and research coverage to more challenging domains.

\vspace{-8pt}

\paragraph{Progressive Multi-round Capability Training}
The first stage progressively establishes the complementary capabilities required for long-horizon research. We first train the model on browse-intensive multi-round trajectories that involve iterative search, webpage reading, evidence acquisition, query reformulation, and final answer synthesis. This phase establishes the model's fundamental tool-use and web-navigation capabilities.
Then we introduce expert-level reasoning data with a greater emphasis on long-form thinking, multi-step deduction, hypothesis verification, and careful answer selection. This phase strengthens the model's ability to solve difficult problems and maintain coherent intermediate conclusions over extended reasoning processes. Together, these two phases allow the model to acquire both iterative agentic behavior and strong reasoning capabilities in a progressive manner.

Although the reasoning-intensive phase strengthens the model's problem-solving ability, we observe that excessive specialization on expert reasoning data may weaken previously acquired browsing and tool-use behaviors. To mitigate this capability interference, we introduce a mixed-capability consolidation stage. In this stage, we combine capability-expanding tasks, including complex academic paper research and expert-level knowledge-intensive reasoning, with selective replay from browse-intensive long-horizon trajectories. Rather than replaying complete trajectories uniformly, we concentrate the replay objective on difficult and informative intermediate decisions. The identification and targeted optimization of these steps are described in the following paragraph.

\vspace{-8pt}

\paragraph{Key-step Focused Mixed-Capability Consolidation.}
\label{subsec: key-step}
A key challenge in long-horizon research training is that supervision is highly uneven across trajectory steps. Most steps are routine transitions, whereas a small number correspond to critical research decisions, such as discovering answer-relevant evidence, redirecting search after rejecting an incorrect hypothesis, or connecting distributed evidence to the target question. These steps often determine whether the trajectory succeeds.

Full-trajectory training applies supervision to all assistant tokens in a successful trajectory, which can dilute the learning signal from high-value decision points. To address this issue, we introduce key-step focused supervision. We first identify candidate key steps using high-precision rule-based detectors that target semantically meaningful research events, such as:
\begin{itemize}
\item the first tool invocation whose results or webpage observations provide evidence for an answer-relevant entity or constraint after multiple exploratory steps;
\item the first step where the model rejects previously explored incorrect entities or hypotheses and redirects the search toward a more promising direction;
\item key context-update steps,where the model invokes the context-update tool to preserve verified evidence in the improvement state, record unresolved conditions, and prepare the next targeted research plan. 
\end{itemize}

These annotations are grounded in verifiable task structure rather than the model's self-reported reasoning. 
A step is marked as critical only when a valid tool observation introduces answer-relevant evidence, supports a verified transition toward the solution, or produces an actionable improvement state after evidence accumulation. 
We retain key-step annotations only for trajectories that pass final-answer verification; merely mentioning an answer-related entity, repeating tool calls, or claiming progress is therefore insufficient without externally grounded evidence and a verified successful outcome. All annotations are constructed offline for training tasks with verifiable reference answers and are not used during evaluation or inference.

To verify whether these detected steps are indeed difficult for the model to learn, we perform a step-level loss analysis after full-trajectory training. For each assistant step $s_j$ in a trajectory, we compute the average token loss:
$$
\ell(s_j) =
-\frac{1}{|s_j|}
\sum_{k=1}^{|s_j|}
\log \pi_{\theta}(a_{j,k} \mid c_{j,k}),
$$
where $a_{j,k}$ denotes the $k$-th assistant token in step $s_j$, and $c_{j,k}$ is its preceding context. We find that the detected key steps consistently have substantially higher loss than ordinary trajectory steps. This suggests that even after full-trajectory supervision, the model still underfits the most informative and decision-critical parts of the trajectory.

Based on this observation, we construct a key-step training set. For each selected key step, we preserve its full preceding context so that the model is trained under the correct trajectory state. However, the loss is applied only to the key step itself, while the prefix tokens, user messages, and tool observations are masked out. The key-step objective is:
$$
\mathcal{L}_{\mathrm{key}}
=
-\mathbb{E}_{s_j \sim \mathcal{K}}
\left[
\frac{1}{|s_j|}
\sum_{k=1}^{|s_j|}
\log \pi_{\theta}(a_{j,k} \mid c_{j,k})
\right],
$$
where $\mathcal{K}$ denotes the set of selected key steps.

This design preserves the conditioning context of the selected step while concentrating the optimization signal on high-value research decisions. Empirically, adding key-step focused supervision after full-trajectory training further improves the model's performance on long-horizon deep research tasks, indicating that selective supervision over informative decision points can improve the effectiveness of selective agentic mid-training. 

\subsection{Step-aware Reinforcement Learning}
\paragraph{Step-aware Group Policy Optimization}
Our mid-training analysis shows that long-horizon research trajectories are highly heterogeneous: routine tool-use or transition steps are often easy to imitate, while a small number of decisive steps are much harder and more important for final task success. 
This observation also motivates our reinforcement learning design. In deep research tasks, a trajectory may contain many rounds of search, browsing, hypothesis update, and evidence verification. Since external tools can return noisy or partially relevant observations, not every action in a successful trajectory should receive the same learning signal.

Standard group-relative policy optimization methods compute a sequence-level reward and propagate the resulting advantage to all generated tokens in the trajectory. 
However, a sequence-level advantage provides relatively coarse credit assignment for long-horizon tool-use trajectories, where different assistant steps may serve substantially different roles. 
We adopt a turn-level policy optimization formulation and adapt it to long-horizon deep research through hierarchical step-balanced normalization and key-step shaping signals shared with the mid-training stage.

For each prompt $x$, we sample $G$ trajectories from the old policy. Let $M_i$ denote the number of assistant steps in the $i$-th trajectory, and let $L_{i,j}$ denote the number of generated tokens in its $j$-th step. 
We construct a step-specific shaped advantage $A_{i,j}$ for each assistant step. All steps in the same trajectory share the group-relative outcome advantage, while annotated key steps receive an additional auxiliary bonus.

For token $a_{i,j,k}$, we define the token-level probability ratio as
$$
r_{i,j,k}(\theta) =
\frac{
\pi_{\theta}(a_{i,j,k}\mid c_{i,j,k})
}{
\pi_{\theta_{\mathrm{old}}}(a_{i,j,k}\mid c_{i,j,k})
},
$$
where $c_{i,j,k}$ is the context before token $a_{i,j,k}$. We then aggregate token-level ratios into a length-normalized step-level policy ratio using the geometric mean:
$$
\rho_{i,j}(\theta) = 
\exp
\left(
\frac{1}{L_{i,j}}
\sum_{k=1}^{L_{i,j}}
\log r_{i,j,k}(\theta)
\right).
$$
The length normalization keeps the policy-ratio scale comparable across assistant steps of different lengths. 
We then apply hierarchical averaging over steps and trajectories to prevent longer trajectories from dominating the training objective.

We adopt hierarchical normalization by first averaging over assistant steps within each trajectory and then over trajectories within the rollout group:
$$
\mathcal{L}_{\mathrm{step}}
=
-
\mathbb{E}_{x \sim \mathcal{D}}
\left[
\frac{1}{G}
\sum_{i=1}^{G}
\frac{1}{M_i}
\sum_{j=1}^{M_i}
\min
\left(
\rho_{i,j}(\theta) A_{i,j},
\operatorname{clip}
\left(
\rho_{i,j}(\theta),
1-\epsilon,
1+\epsilon
\right)
A_{i,j}
\right)
\right].
$$
This prevents trajectories with more interaction steps from dominating the objective solely because of their length and better reflects the heterogeneous structure of long-horizon research trajectories.

We add a KL penalty against a reference policy:
$$
\mathcal{L}
=
\mathcal{L}_{\mathrm{step}}
+
\beta_{\mathrm{KL}}
\mathbb{E}
\left[
D_{\mathrm{KL}}
\left(
\pi_{\theta}(\cdot\mid c)
\Vert
\pi_{\mathrm{ref}}(\cdot\mid c)
\right)
\right].
$$
\paragraph{Step Reward Shaping}
In addition to outcome rewards, we introduce step-level reward shaping to emphasize informative intermediate decisions. 
This design follows the key observation from mid-training: some steps in a long research trajectory are more decisive than others. 
The key-step annotations described in Section~\ref{sec:mid-train} are reused as bounded auxiliary shaping signals.

Let $R_i$ denote the trajectory-level outcome reward for $\tau_i$. We first compute the group-relative outcome advantage:
$$
A^{\mathrm{out}}_i =
\frac{
R_i - \mu_R
}{
\sigma_R + \epsilon
},
$$
where $\mu_R$ and $\sigma_R$ are the mean and standard deviation of rewards within the sampled group.

For each assistant step $s_{i,j}$, we define a key-step indicator $B_{i,j}$ based on the key-step annotations described in Section~\ref{subsec: key-step}. 
To avoid rewarding spurious intermediate behavior, the key-step bonus is applied only when the trajectory-level result is valid:
$$
\widetilde{B}_{i,j} = 
\mathbb{I}[R_i > 0]\cdot B_{i,j}.
$$

We directly add a bounded auxiliary bonus to annotated key steps in successful trajectories. 
The final step-level advantage is defined as:
$$
A_{i,j}
=
A^{\mathrm{out}}_{i}
+
\lambda_{\mathrm{key}}
\widetilde{B}_{i,j},
$$
where $\lambda_{\mathrm{key}}$ controls the strength of step-level shaping.

This formulation preserves the final-answer reward as the main optimization signal while adding a small auxiliary preference for decision-critical steps. The shaping term is not intended to solve general credit assignment for all possible intermediate actions. Instead, it provides high-precision supervision for a subset of research steps that are both identifiable and empirically important. We use this lightweight shaping signal as part of the final AREX reinforcement learning recipe, while retaining final-answer correctness as the primary optimization objective. 

\section{Experiments}

\subsection{Experimental Setup}

\paragraph{Benchmarks.}
We evaluate \textsc{AREX} on six benchmarks spanning four complementary regimes of search-augmented reasoning.
\textbf{BrowseComp} and \textbf{DeepSearchQA} emphasize deep research, requiring multi-step retrieval, query reformulation, evidence aggregation, and answer synthesis. 
\textbf{GAIA} and \textbf{xbench-2510} additionally stress agentic task completion, where information seeking must be coordinated with planning, tool use, and multi-step reasoning. 
\textbf{WideSearch} measures broad-coverage retrieval and synthesis over a large search space; we report results on its English subset. Finally, \textbf{HLE with tools} evaluates high-level reasoning with access to web search and computational tools. We report the Item-F1 score for \textbf{WideSearch}, F1 score for \textbf{DeepSearchQA} and accuracy for all other benchmarks. 

\vspace{-8pt}

\paragraph{Evaluation protocols.}
We evaluate our models with a unified long-horizon search-agent interface comprising \texttt{search}, \texttt{visit}, \texttt{update context}, and \texttt{finish} tools. For \textbf{HLE with tools}, we add a \texttt{python} tool. 
The agent can iteratively retrieve webpages, inspect sources, maintain a compact working state, and return a structured final response. Each episode allows at most 300 \emph{inner research loop turns} and 5 \emph{outer self-improvement loop} operations following~\cite{team2026mirothinker}.

\subsection{Overall Performance}

\begin{table*}[t]
   \vspace{-3pt}
    \centering
    \scriptsize
    \renewcommand{\arraystretch}{1.12}
    \setlength{\tabcolsep}{3.2pt}
    \resizebox{\textwidth}{!}{%
    \begin{tabular}{@{}lrrrrrr@{\hspace{0.8em}}}
    \toprule
    \textbf{Model} 
    & \textbf{BrowseComp} 
    & \textbf{GAIA} 
    & \textbf{xbench-2510} 
    & \textbf{DeepSearchQA} 
    & \textbf{WideSearch-en} 
    & \textbf{HLE (tool)} \\
    \midrule

    \rowcolor{gray!10}
    \multicolumn{7}{@{}l}{\textbf{Frontier models}} \\
    GPT-5.4 & 82.7 & -- & -- & 88.5 & 77.5 & 52.1\rlap{\textsuperscript{*}} \\
    Opus-4.6 & 83.7 & -- & -- & 91.3 & 77.5 & 53.0\rlap{\textsuperscript{*}} \\
    Gemini-3.1-Pro & 85.9 & 80.6 & 53.0 & 93.3 & 66.4 & 51.4\rlap{\textsuperscript{*}} \\
    
    \midrule
    \rowcolor{gray!10}
    \multicolumn{7}{@{}l}{\textbf{Open-source models}} \\
    GLM-5 & 75.9 & 70.0 & -- & -- & 69.8 & 50.4 \\
    Kimi-K2.6 & 83.2 & 80.6 & 90.0 & 92.5 & 80.8 & 54.0\rlap{\textsuperscript{*}} \\
    DeepSeek-V4-Flash & 73.2 & -- & 69.0 & 90.6 & 76.4 & 45.1 \\
    DeepSeek-V4-Pro & 83.4 & -- & 80.0 & 88.7 & 78.0 & 48.2 \\
    Tongyi-DeepResearch-30B & 43.4 & 70.9 & 55.0 & -- & -- & 32.9 \\
    Qwen3.5-35B & 61.0 & 80.0 & 50.3 & 68.5 & 57.1 & 47.4 \\
    Qwen3.5-122B & 63.8 & 81.6 & -- & -- & 60.5 & 47.5 \\
    Qwen3.5-397B & 78.6 & 83.5 & 61.0 & 82.1 & 74.0 & 48.3 \\
    MiroThinker-1.7-mini & 67.9 & 80.3 & 57.2 & 67.9 & -- & 36.4 \\
    MiroThinker-1.7 & 74.0 & 82.7 & 62.0 & 72.1 & -- & 42.9 \\
    MiroThinker-H1 & 88.2 & 88.5 & 72.0 & 80.6 & -- & 47.7 \\
    Quest-35B & 64.6 & 80.8 & -- & -- & 60.6 & 37.2 \\

    \midrule
    \rowcolor{blue!7}
    \multicolumn{7}{@{}l}{\textbf{Ours}} \\
    \rowcolor{blue!3}
    \textsc{AREX}-Turbo & 70.7 & 81.6 & 57.0 & 78.5 & 68.5 & 40.6 \\
    \rowcolor{blue!3}
    \textsc{AREX}-Base & 82.5 & 85.4 & 71.0 & 89.9 & 82.0 & 52.4 \\
    
    \bottomrule
    \end{tabular}%
    }
    \caption{Comparison results on deep search and agentic reasoning tasks. Models are grouped into frontier closed-source models, open-source models, and our \textsc{AREX} models. $^\ast$ denotes results reported on the full HLE; unmarked results are reported on the text-only subset.}
    \label{tab:benchmark-comparison}
\end{table*}

As shown in Table~\ref{tab:benchmark-comparison}, \textsc{AREX} achieves strong and consistent performance across diverse search-augmented reasoning tasks. 
When accounting for model scale, \textsc{AREX}-Base demonstrates a strong capability-to-parameter trade-off with 10B active parameters.
It consistently improves over the Qwen3.5 backbone family, including the substantially larger Qwen3.5-397B model, and remains competitive with leading open-source and proprietary research agents across all evaluations. 
Beyond the Qwen3.5 family, \textsc{AREX}-Base also compares favorably with specialized research agents and other frontier open-source systems. It outperforms MiroThinker-H1 on DeepSearchQA and text-only HLE while remaining within one point on xbench-2510; it also surpasses DeepSeek-V4-Pro on DeepSearchQA, WideSearch-en, and text-only HLE, and exceeds Kimi-K2.6 on GAIA and WideSearch-en. Against proprietary frontier systems, \textsc{AREX}-Base achieves the best reported WideSearch-en score and remains competitive on both BrowseComp and DeepSearchQA.
These gains extend beyond search-intensive tasks to broad information synthesis, long-horizon agentic tasks, and expert-level reasoning, with \textsc{AREX}-Base achieving the best overall result on WideSearch-en. 
A similar trend holds at the compact scale, where the 4B \textsc{AREX}-Turbo outperforms Qwen3.5-35B on five of the six benchmarks. 
Taken together, these results demonstrate that \textsc{AREX} possesses strong and broadly transferable research capabilities across diverse search, reasoning, and tool-use settings. 
Its favorable performance at both model scales further demonstrates the effectiveness of our framework, which recursively preserves verified progress and redirects search toward unresolved constraints to enable systematic progress over long research horizons.

\subsection{Inference Framework Analysis}

At inference time, \textsc{AREX} executes the inner research loop under a fixed interaction budget. 
The model uses \texttt{python}, \texttt{search}, and \texttt{visit} to acquire and verify evidence, \texttt{update\_context} to consolidate the evolving research trajectory into a compact research state, and \texttt{finish} to externalize an answer together with its supporting evidence and confidence score. 
We analyze two mechanisms that determine how \textsc{AREX} uses its inference budget: Autonomous Context Updating (ACU), which refreshes the effective context inside the inner research loop, and the outer self-improvement loop, which accepts, refines, or restarts a recursive round based on the structured result produced by \texttt{finish}. 
The analysis separates three questions: when the model invokes \texttt{update\_context}, how ACU and the outer self-improvement loop change accuracy under controlled comparisons, and whether the answer-level confidence score supports the confidence-based decision procedure. 
We refer to the complete system as \textsc{AREX} w/ ACU. In the corresponding ablation, denoted as \textsc{AREX} w/o ACU, the same model retains all other tools and operates under the same interaction budget, but cannot refresh its research state and must instead condition on the uncompressed interaction history.

\vspace{-8pt}

\paragraph{Context-Update Behavior.}
We first examine ACU as a runtime behavior rather than only as an architectural component. The agent operates within a 128K-token active context window. 
Within this budget, \textsc{AREX} may invoke \texttt{update\_context} when the current research trajectory needs consolidation. If the active context reaches the 128K-token limit, the agent must invoke \texttt{update\_context} before continuing its research.

\begin{table*}[t]
\centering
\small
\setlength{\tabcolsep}{4.5pt}
\renewcommand{\arraystretch}{1.15}
\begin{tabular*}{\textwidth}{@{\extracolsep{\fill}}lrlrlr@{}}
\toprule
\multicolumn{2}{c}{\textbf{Update usage}} &
\multicolumn{2}{c}{\textbf{Call trigger}} &
\multicolumn{2}{c}{\textbf{Update content}} \\
\cmidrule(lr){1-2}\cmidrule(lr){3-4}\cmidrule(lr){5-6}
\textbf{Metric} & \textbf{Value} &
\textbf{Category} & \textbf{Share} &
\textbf{Category} & \textbf{Share} \\
\midrule
Cases w/ update & 80.3\% & Revise search strategy & 66.9\% & Verified findings & 72.1\% \\
Upper bound & 128{,}000 & Reject candidate & 13.6\% & Current candidates & 39.2\% \\
Min.\ context tokens & 3{,}788 & Verify evidence/answer & 5.3\% & Unresolved constraints & 95.5\% \\
Mean context tokens & 25{,}721 & Identify new lead & 7.2\% & Validity concerns & 14.1\% \\
Median context tokens & 25{,}386 & Summarize progress & 6.4\% & Rejected candidates & 81.5\% \\
Max.\ context tokens & 128{,}591 & Other & 0.6\% & Next-step plan & 96.4\% \\
\bottomrule
\end{tabular*}
\caption{Context-update behavior on BrowseComp.
Context-token statistics are measured when \texttt{update\_context} is invoked.
Trigger shares are single-label, whereas content shares are multi-label.}
\label{tab:browsecomp-context-management}
\end{table*}

Table~\ref{tab:browsecomp-context-management} shows that \textsc{AREX} invokes \texttt{update\_context} in 80.3\% of BrowseComp cases. Updates occur at a mean active-context size of 25,721 tokens and a median of 25,386 tokens, well below the configured 128K-token upper bound. Only 0.01\% of updates occur at or above the limit. This timing indicates that ACU is used mainly as a proactive research operation, while the hard limit acts as a safety constraint rather than the dominant trigger.

The call triggers show where the research policy decides that a compact state is needed. Search-strategy revision accounts for 66.9\% of calls, followed by candidate rejection at 13.6\%. Identifying new leads, summarizing progress, and verifying evidence or a tentative answer account for smaller shares. The model most often refreshes its state when the retrieval direction has become unproductive or when new evidence changes which candidates remain viable.

The update contents highlight the importance of ACU in preserving decision-relevant research state. Overall, ACU retains most of the effective information needed for subsequent reasoning, especially unresolved constraints and next-step plans, which are the most consistently preserved elements. This suggests that ACU prioritizes information about what remains to be checked, what conditions still need to be satisfied, and what actions should be taken next. It also retains verified findings and rejected candidates, ensuring that later decisions are grounded in established evidence while avoiding previously invalidated paths. In this sense, ACU functions as a compact representation of the research process, organizing it around confirmed evidence, remaining constraints, eliminated alternatives, and planned follow-up actions.

Together, the usage, trigger, and content patterns characterize ACU as a trajectory consolidation and research-state refreshing mechanism for long-horizon research. By updating at meaningful turning points and preserving both positive and negative evidence, ACU helps \textsc{AREX} avoid previously rejected paths and spend the remaining interaction budget on unresolved constraints.

\vspace{-8pt}

\paragraph{Recursive Self-Improvement.}

\begin{figure*}[t]
\centering
\begin{minipage}[c]{0.33\linewidth}
\centering
\renewcommand{\arraystretch}{1.18}
\setlength{\tabcolsep}{3.5pt}
\small
\begin{tabular}{@{}llc@{}}
\toprule
\textbf{Method} & \textbf{Setting} & \textbf{Acc.} \\
\midrule
\multirow{2}{*}{\textit{\textsc{AREX} w/o ACU}}
& w/o outer loop & 59.6 \\
& w/ outer loop  & 69.8 \\
\addlinespace
\multirow{2}{*}{\textit{\textsc{AREX} w/ ACU}}
& w/o outer loop & 71.4 \\
& w/ outer loop  & 82.5 \\
\bottomrule
\end{tabular}
\captionof{table}{Effect of ACU in the inner research loop and the outer self-improvement loop. Accuracy is measured on BrowseComp.}
\label{tab:browsecomp-accuracy}
\end{minipage}
\hfill
\begin{minipage}[c]{0.64\linewidth}
\centering
\includegraphics[width=\linewidth]{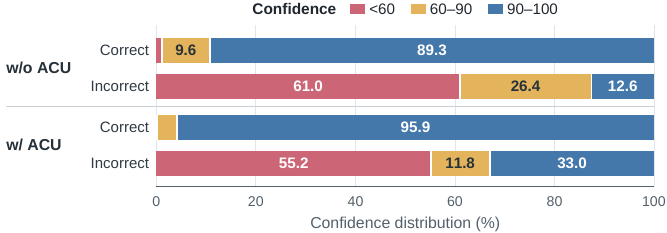}
\vspace{-2em}
\captionof{figure}{Confidence distributions for correct and incorrect outputs. Each bar is normalized within its method and outcome group.}
\label{fig:browsecomp-confidence}
\end{minipage}
\vspace{-0.5em}
\end{figure*}

Table~\ref{tab:browsecomp-accuracy} separates the two parts of recursive self-improvement: ACU in the inner research loop and the outer self-improvement loop. The rows without the outer loop accept the first output produced by \texttt{finish}, so they isolate the effect of replacing the complete trajectory with the effective context constructed by ACU. Under this matched single-round setting, ACU raises BrowseComp accuracy from 59.6 to 71.4, an absolute gain of 11.8 points. The gain supports the inner-loop claim that a refreshed research state containing verified findings, rejected candidates, unresolved constraints, and source identifiers is more useful to the research policy than the complete interaction history.

The within-method differences isolate the effect of the outer self-improvement loop after an inner loop has produced a structured result. Without ACU, enabling the outer loop increases accuracy from 59.6 to 69.8, an absolute gain of 10.2 points. With ACU, the same outer loop raises accuracy from 71.4 to 82.5, a gain of 11.1 points. These consistent gains show the importance of the outer self-improvement loop: \textsc{AREX} improves when it can run another recursive round after a low-confidence structured result rather than returning the first answer produced by \texttt{finish}. The full system reaches 82.5 accuracy, 22.9 points above the configuration with neither ACU nor the outer self-improvement loop.

\vspace{-8pt}

\paragraph{Answer-Level Confidence Score.}

Figure~\ref{fig:browsecomp-confidence} evaluates the confidence score $s^{(k)}$ produced by \texttt{finish}, which is part of the structured result passed from the inner loop to the outer self-improvement loop. Correct final outputs concentrate in the 90--100 bin, with 89.3\% of correct outputs in this range without ACU and 95.9\% with ACU. Incorrect final outputs retain a large low-confidence mass: 61.0\% of errors without ACU and 55.2\% with ACU fall below 60. This separation supports the confidence-based decision procedure because many failures can be identified from the answer-level confidence score without reinterpreting the entire trajectory.

\subsection{Ablation Studies}

We conduct controlled ablations on BrowseComp to examine three components of the \textsc{AREX} training recipe: progressive multi-round capability training, key-step focused supervision, and step-aware reinforcement learning. All variants use the same model initialization and retain the remaining training pipeline. For supervision-related ablations, we match the supervised-token and optimization budgets; for reinforcement learning, we keep the prompts, rollout budget, initialization, and training schedule unchanged.

\vspace{-8pt}

\paragraph{Progressive multi-round capability training.}
The first stage of our mid-training pipeline is designed to acquire complementary capabilities in sequence. The model is first trained on browse-intensive, multi-round tool-interaction trajectories to establish web navigation, evidence acquisition, and iterative research behaviors. It is then trained on reasoning-intensive trajectories that place greater emphasis on long-form reasoning, hypothesis verification, and difficult problem solving. To evaluate whether this progressive ordering is necessary, we replace the progressive multi-round capability-training procedure with direct mixed training, where the browse-intensive and reasoning-intensive data are combined from the beginning under a matched overall training budget. This ablation tests whether progressively acquiring tool-use and reasoning capabilities is more effective than learning them simultaneously from a single mixed distribution.

\begin{figure*}[t]
    \centering

    \begin{minipage}[t]{0.45\textwidth}
        \vspace{0pt}
        \centering
        \includegraphics[width=0.83\linewidth]{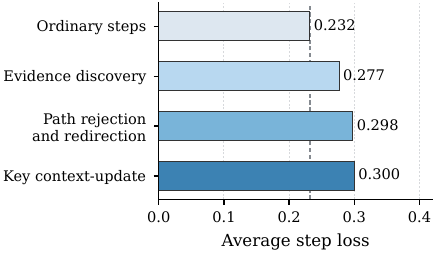}
        \vspace{-0.7em}
        \captionsetup{type=figure}
        \captionof{figure}{
            Average step loss after full-trajectory mid-training.
        }
        \label{fig:key-step-loss}
    \end{minipage}
    \hfill
    \begin{minipage}[t]{0.54\textwidth}
        \vspace{0pt}
        \centering
        \resizebox{\textwidth}{!}{
        \small
        \renewcommand{\arraystretch}{1.5}
        \setlength{\tabcolsep}{7pt}
        \begin{tabular}{@{}lc@{}}
            \toprule
            \textbf{Training setting} & \textbf{Accuracy} \\
            \midrule
            Replace multi-round capability training by mixed training
                & 77.5 \\
            Replace key-step focused supervision by random-step replay
                & 74.1 \\
            Replace step-aware RL by standard GRPO
                & 79.4 \\
            Full \textsc{AREX}
                & 82.5 \\
            \bottomrule
        \end{tabular}
        }
        \vspace{3pt}
        \captionsetup{type=table}
        \captionof{table}{
            Ablation of the AREX training recipe.
        }
        \label{tab:arex-key-step-ablation}
    \end{minipage}
\end{figure*}

As shown in Table~\ref{tab:arex-key-step-ablation}, replacing progressive multi-round capability training with direct mixed training reduces BrowseComp accuracy from 82.5 to 77.5. Although both settings use the same browse-intensive and reasoning-intensive capability distributions, the staged recipe first establishes multi-round tool-use behavior before introducing reasoning-heavy supervision. The result suggests that this ordering reduces interference between heterogeneous training objectives and provides a stronger initialization for the subsequent consolidation and reinforcement learning stages.

\vspace{-8pt}

\paragraph{Step-level loss analysis.}
We analyze whether the key steps identified in Section~\ref{sec:mid-train} remain difficult after full-trajectory mid-training. For each assistant step, we compute its average token-level negative log-likelihood under the trained model and compare the average losses of annotated key steps and ordinary steps.

As illustrated in Figure~\ref{fig:key-step-loss}, ordinary steps have an average loss of 0.232, whereas all three annotated key-step categories exhibit higher loss. Evidence discovery, path rejection and redirection, and key context-update obtain average losses of 0.277, 0.298, and 0.300, respectively. These values correspond to relative increases of approximately 19\%, 28\%, and 29\% over ordinary steps. The consistent gap across all three categories suggests that full-trajectory supervision learns routine actions more readily, while intermediate decisions that establish evidence, revise the research direction, or resolve context updates remain comparatively underlearned.

\vspace{-8pt}

\paragraph{Key-step focused supervision and step-aware reinforcement learning.}
We evaluate key-step focused supervision and step-aware reinforcement learning on BrowseComp under matched training conditions. For the variant without key-step focused supervision, selected key steps are replaced with randomly sampled assistant steps from the same browse-intensive trajectory pool. The random-step baseline uses the same prefix-conditioned training format, applies loss only to the sampled step, and matches the supervised-token and optimization budgets. For the variant without step-aware reinforcement learning, we replace the complete step-aware objective with standard GRPO~\citep{shao2024deepseekmath, schulman2017proximal} while keeping the RL prompts, rollout budget, initialization, and training schedule unchanged.

Replacing key-step focused supervision with equal-budget random-step replay reduces BrowseComp accuracy from 82.5 to 74.1, representing the largest degradation among the ablations. Because ordinary steps constitute the majority of long-horizon trajectories, the random-step baseline predominantly replays routine behaviors that already exhibit relatively low loss. Together with Figure~\ref{fig:key-step-loss}, this result indicates that additional supervision is substantially more effective when concentrated on underlearned, decision-critical steps than when assigned to arbitrary intermediate actions.

Finally, replacing the step-aware reinforcement learning objective with standard GRPO lowers accuracy from 82.5 to 79.4 under the same rollout and training configuration. The resulting 3.1-point gain suggests that step-aware policy optimization further refines the research policy after the main capabilities have been established during mid-training.

Overall, the ablations show that the performance of \textsc{AREX} is supported by complementary training components. Progressive multi-round capability training establishes long-horizon interaction behavior, key-step focused supervision reinforces difficult and decisive intermediate actions, and step-aware reinforcement learning further refines the resulting research policy.

\section{Conclusion}
\label{sec:conclusion}

In this work, we introduced \textsc{AREX}, a recursively self-improving agent for deep research. 
\textsc{AREX} exploits the discovery--verification asymmetry in deep research to transform provisional answers into partially verified research states, preserving supported evidence while isolating unresolved claims for subsequent targeted investigation. 
Coupled with autonomous context updating and step-aware training, \textsc{AREX} achieves strong performance across deep search, wide search, agentic reasoning, and tool-use benchmarks. 
Our preliminary training analysis further highlights the importance of identifying and reinforcing decision-critical steps in long-horizon trajectories, rather than treating all intermediate actions uniformly.
Future work will investigate more general and autonomous mechanisms for estimating step utility and assigning fine-grained training signals.
These results suggest that verification-guided state refinement and step-aware optimization offer a promising direction for building reliable long-horizon research agents.



\bibliographystyle{assets/plainnat}
\bibliography{citation}
\clearpage

\appendix

\section{Related Work}

\paragraph{Tool-Augmented Deep Research.}
Language-model agents extend LLMs from static response generation to iterative problem solving with tools, web environments, and external knowledge sources~\citep{react,toolformer,webgpt}. Deep research pushes this paradigm further: agents must locate sparse evidence, compare conflicting sources, and synthesize answers that satisfy multiple coupled constraints~\citep{deepresearch,deepresearch_survey}. Recent systems improve research performance through tool-use supervision, synthetic trajectory construction, agentic fine-tuning, reinforcement learning, and inference-time scaling over longer search traces~\citep{searchr1,webdancer,websailor,tongyi_deepresearch,minddr,beyond_ten_turns}. These approaches primarily strengthen exploration within a research trajectory. \textsc{AREX} addresses a complementary problem: after partial progress has been made, the agent must identify which constraints are already supported, which remain unresolved, and how this diagnosis should define the next research problem.

\paragraph{Verification as Recursive Research Control.}
Verification has been widely used to improve reasoning, typically as outcome ranking, process supervision, or step-level reward modeling~\citep{lets_verify,math_shepherd}. 
Recent work further exploits the asymmetry between generating a correct solution and verifying a proposed one, using verification to guide test-time search or critique intermediate agent decisions~\citep{zeng2025pushing,team2026mirothinker}.
\textsc{AREX} uses verification in a different role. Rather than treating it as a final acceptance filter or a local action critic, \textsc{AREX} makes verification the transition operator between research rounds. Constraint-wise auditing converts a provisional answer into a partially verified research state: supported claims are preserved, unresolved conditions are isolated, and contradictory evidence is surfaced. The next round of research is therefore driven by the structure of remaining uncertainty, not by undirected continuation of search.

\paragraph{Research-State Management and Long-Horizon Training.}
Sustained research requires agents to maintain useful information across long trajectories containing evidence, failed searches, speculative hypotheses, and evolving plans. Prior work studies hierarchical memory, virtual context management, learned memory states, periodic summarization, and explicit memory-editing actions~\citep{memgpt,hiagent,mem1,resum,supo,memory_as_action,xie2026quest}. \textsc{AREX} builds on this direction by learning to invoke a context-update tool that produces an improvement state organized around the current research objective, including verified evidence, citations, constraint status, unresolved gaps, and the next plan. Training such agents also raises sparse credit-assignment challenges, since final success rarely reveals which intermediate steps were decisive. Inspired by process supervision and turn-level credit assignment~\citep{lets_verify,math_shepherd,ragen,turn_credit}, \textsc{AREX} increases exposure to key steps where evidence is acquired, contradictions are resolved, or incorrect directions are repaired, enabling recursive research control at efficient dense and Mixture-of-Experts model scales~\citep{switch_transformer,gshard,mixtral,deepseekv3}.

\section{Preliminary Exploration under a Simplified Setting}

This section investigates a training direction that was not incorporated into the final \textsc{AREX}-Base recipe. The experiment uses an early simplified configuration with a reduced tool set and without ACU, outer self-improvement loop, key-step supervision, the complete multi-stage data mixture, or the full test-time scaling strategy. Therefore, its absolute BrowseComp scores are not directly comparable with either the complete \textsc{AREX}-Base system or results reported for the original backbone. The comparison is intended only to measure relative differences within the same controlled setting.

\paragraph{Trajectory self-distillation.}
We conduct a preliminary experiment to examine whether a browse-trained agent can generate improved supervision for a fresh model initialized from the same backbone. Both the direct-training and self-distillation variants use the 122B-A10B initialization, the same problem distribution, and matched training and inference configurations. The direct-training baseline learns from the original browse-intensive trajectories. For self-distillation, an intermediate browse-trained agent regenerates trajectories for the same problems, after which a fresh copy of the base model is trained on the accepted generated trajectories.

Under this simplified setting, trajectory self-distillation improves BrowseComp from 52.3 to 57.1, a gain of 4.8 points (Table~\ref{tab:arex-self-distillation}). This result suggests that trajectories generated by an intermediate agent may provide more effective supervision than the original trajectories for the same backbone. Possible explanations include better alignment with the target policy's action distribution, more consistent search decisions, and cleaner intermediate reasoning structure.

This preliminary study does not isolate the source of the improvement or establish that the gain persists when combined with ACU, key-step supervision, and the complete \textsc{AREX} training recipe. Moreover, self-generated trajectories may inherit or amplify the teacher model's biases and failure modes. We therefore present trajectory self-distillation as a promising direction for future iterations rather than as a component of the final system.

\begin{table*}[t]
    \centering
        \vspace{0pt}
        \centering
        \captionsetup{type=table}
        \captionof{table}{
            Exploratory self-distillation experiments on BrowseComp. Both configurations use the 122B-A10B backbone without ACU, outer self-improvement loop, or key-step supervision.
        }
        \label{tab:arex-self-distillation}

        \small
        \renewcommand{\arraystretch}{1.15}
        \setlength{\tabcolsep}{7pt}

        \begin{tabular}{@{}lc@{}}
            \toprule
            \textbf{Training setting} & \textbf{BrowseComp} \\
            \midrule
            Direct training   & 52.3 \\
            Self-distillation & 57.1 \\
            \bottomrule
        \end{tabular}
    \hfill
   
\end{table*}

\clearpage

\section*{Contributions}

\begingroup
\setlength{\parindent}{0pt}
\setlength{\parskip}{0pt}
\raggedright

{\color{contributionblue}

\textbf{Core Contributors:}
Shuqi Lu\footnote{These authors contributed equally to this work.}, \
Chaofan Li\footnotemark[1], \
Kun Luo\footnotemark[1], \
Zhang Zhang, \ 
Hui Wang, \ 
Hongwang Xiao, \ 
Zheng Liu\footnote{Zheng Liu is the project leader.}

\vspace{0.5em}

\textbf{Participants:}
Lei Xiong, \ 
Jiahao Wang, \ 
Sen Wang, \ 
Xiyan Jiang, \ 
Wanli Li, \ 
Yuyang Hu, \ 
Hongjin Qian, \ 
Bingyu Yan, \ 
Jianlyu Chen, \ 
Ziyi Xia \ 

\vspace{0.5em}

\textbf{Advisors:}
Yingxia Shao, \ 
Kang Liu, \ 
Zhicheng Dou, \ 
Di He, \ 
Chaozhuo Li, \ 
Qiwei Ye, \ 
Zhongyuan Wang 



\vspace{1.5em}


}















\endgroup

\end{document}